RESEARCH ARTICLE

# RDF-star2Vec: RDF-star Graph Embeddings for Data Mining

SHUSAKU EGAMI[1], TAKANORI UGAI[1,2], MASATERU OOTA[1], KYOUMOTO MATSUSHITA[2], TAKAHIRO KAWAMURA[3], KOUJI KOZAKI[4], AND KEN FUKUDA[1]
[1]National Institute of Advanced Industrial Science and Technology, Koto, Tokyo 135-0064, Japan
[2]Fujitsu Ltd., Kawasaki, Kanagawa 105-7123, Japan
[3]National Agriculture and Food Research Organization, Tsukuba, Ibaraki 305-0856, Japan
[4]Department of Engineering Informatics, Faculty of Information and Communication Engineering, Osaka Electro-Communication University, Neyagawa, Osaka 572-0833, Japan

Corresponding author: Ken Fukuda (ken.fukuda@aist.go.jp)

This work was supported in part by the New Energy and Industrial Technology Development Organization (NEDO) under Project JPNP20006 and Project JPNP180013; and in part by the Japan Society for the Promotion of Science (JSPS) KAKENHI under Grant JP19H04168, Grant JP22K18008, and Grant JP23H03688.

**ABSTRACT** Knowledge Graphs (KGs) such as Resource Description Framework (RDF) data represent relationships between various entities through the structure of triples (<*subject*, *predicate*, *object*>). Knowledge graph embedding (KGE) is crucial in machine learning applications, specifically in node classification and link prediction tasks. KGE remains a vital research topic within the semantic web community. RDF-star introduces the concept of a *quoted triple* (QT), a specific form of triple employed either as the subject or object within another triple. Moreover, RDF-star permits a QT to act as compositional entities within another QT, thereby enabling the representation of recursive, hyper-relational KGs with nested structures. However, existing KGE models fail to adequately learn the semantics of QTs and entities, primarily because they do not account for RDF-star graphs containing multi-leveled nested QTs and QT–QT relationships. This study introduces RDF-star2Vec, a novel KGE model specifically designed for RDF-star graphs. RDF-star2Vec introduces graph walk techniques that enable probabilistic transitions between a QT and its compositional entities. Feature vectors for QTs, entities, and relations are derived from generated sequences through the structured skip-gram model. Additionally, we provide a dataset and a benchmarking framework for data mining tasks focused on complex RDF-star graphs. Evaluative experiments demonstrated that RDF-star2Vec yielded superior performance compared to recent extensions of RDF2Vec in various tasks including classification, clustering, entity relatedness, and QT similarity.

**INDEX TERMS** Knowledge graph embedding, RDF2Vec, RDF-star, hyper-relational knowledge graphs, N-ary relation, graph walk.

## I. INTRODUCTION

Knowledge graphs (KGs) such as Resource Description Framework (RDF) data represent relationships between various entities in terms of triples (<*subject*, *predicate*, *object*>), facilitating advanced search and reasoning based on semantic relationships. Despite these capabilities, RDF faces the challenge of adequately representing relations beyond the binary. To address this limitation, RDF-star (formerly spelled RDF*) [1] has garnered considerable interest. This extension introduces the *quoted triple* (QT), a specific type of triple that can serve as the subject or object within another triple. To date, RDF-star has been implemented across numerous triplestores.[1] In addition, several use cases[2] [2], [3] and support tools [4], [5], [6], [7] for RDF-star have already been presented. Furthermore, the RDF-star Working Group[3] was established in August 2022 and is working on recommendations for extending

The associate editor coordinating the review of this manuscript and approving it for publication was Giacomo Fiumara.

---

[1]https://w3c.github.io/rdf-star/implementations.html
[2]https://w3c.github.io/rdf-star/UCR/rdf-star-ucr.html
[3]https://www.w3.org/2022/08/rdf-star-wg-charter/

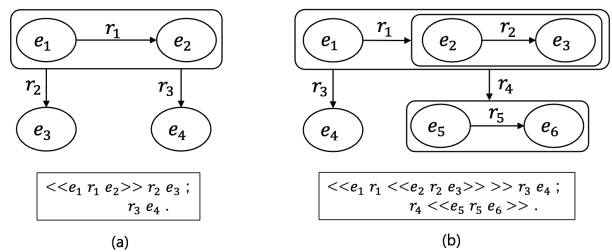

**FIGURE 1.** RDF-star graphs: (a) simple RDF-star and (b) complex RDF-star graphs.

both RDF and SPARQL Protocol and RDF Query Language (SPARQL).[4] Therefore, it is expected that various hyper-relational KGs will be published in RDF-star format in the near future.

Various methods for knowledge graph embedding (KGE) have been proposed in recent years, targeting data mining applications. However, two distinct limitations have been identified in existing KGE methods.

(1) **RDF-star data:** KGE methods for RDF exist [8], [9], but methods specifically tailored for RDF-star data are conspicuously lacking. Although there are approaches to convert RDF-star triples into regular RDF [10] triples, as well as to load non-RDF data formats (e.g., CSV) [11], these techniques fail to capture the vector representations corresponding to QTs, and consequently, compromise the original semantic representation.

(2) **Inability to handle complex RDF-star data:** RDF-star permits the representation of recursive hyper-relational KGs featuring nested structures. This is possible because QT can function as the compositional entities of another QT. However, existing hyper-relational KGE methods [11], [12], [13] can handle only rudimentary structures, as shown in Figure 1(a), and fall short of capturing the semantics of complex structures, including nested QT structures and interrelations between QTs, as shown in Figure 1(b). Additionally, the absence of a publicly available complex RDF-star dataset hinders the further development of RDF-star KGE methods.

This paper introduces RDF-star2Vec, a novel graph walk-based KGE method designed to learn vector representations of normal entities, relations, and QTs in RDF-star graphs with complex structures, including multi-leveled nested QTs and relations between QTs. Specifically, graph walk methods generate sequences that permit probabilistic transition between QTs and asserted triples in RDF-star graphs. Subsequently, feature vectors for QTs, entities, and relations are derived from the generated sequences employing the structured skip-gram model [14]. This approach allows for the direct embedding of RDF-star data into low-dimensional vector space, preserving the semantic representation of the original data. Importantly, the technique situates QTs and their compositional entities in close proximity within the generated sequences, thereby learning the N-ary relations represented by RDF-star. The method was realized through a Java implementation of RDF2Vec [8], [9] and has been released as open-source software.[5]

Furthermore, we introduce a complex RDF-star dataset (KGRC-RDF-star), predicated on KGRC-RDF [15], [16], a scene KG designed for Explainable Artificial Intelligence (XAI) benchmarking and constructed from text data in mystery novels. The dataset features nested statements and scenes such as "Person *A* said 'Person *B* saw Person *C* was in *D*'." Moreover, four gold standard datasets have been constructed to assess classification, clustering, entity relatedness, and QT similarity tasks using the embeddings of KGRC-RDF-star and provided through GEval [17], a KGE evaluation framework. Therefore, this study will augment the existing body of knowledge concerning KGE methods for RDF-star graphs.

The remainder of this paper is structured as follows: Section II introduces related works focusing on KGE methods for RDF, methods for hyper-relational KGs, and benchmarking datasets, and outlines the limitations of these prior works and the position of this paper. Section III introduces our innovative approach, which combines novel graph walks and representation learning methods for RDF-star. Section IV describes the construction of our complex RDF-star graph dataset, featuring multi-leveled nested structures employed in our experimental analysis. Section V describes evaluation tasks, describes the gold standard dataset, and discusses the evaluation results and parameter analysis. Section VI concludes the paper, including a brief summary and future works.

## II. RELATED WORK
### A. KNOWLEDGE GRAPH EMBEDDINGS FOR RDF
Numerous KGE methods have been proposed [18], encompassing graph walk-based, translation-based, and graph neural network methods. This paper focuses on graph walk-based methods applied to RDF graphs. RDF2Vec [8], [9] is a well-known graph walk-based KGE method for RDF graphs. Initially, RDF2Vec generates a sequence set using a random walk. Subsequently, vertices and edges are relabeled employing Weisfeiler-Lehman (WL) graph kernels for RDF [19]. The finalized sequence set serves as input to word2vec [20]. Cochez et al. [21] extended this by incorporating biased random walks for more semantically rich walks. Portisch et al. [22] proposed RDF2Vec Light, a streamlined variant of RDF2Vec that reduces computational complexities by generating vectors only for entities of interest. Moreover, Portisch et al. introduced an order-aware RDF2Vec (RDF2Vec$_{oa}$) [23] using structured word2vec [14], focusing on the original word2vec model's positional insensitivity. Additionally, they proposed similarity-oriented and relevance-oriented walk methods and identified the advantages of each method [24].

---
[4]SPARQL is a recursive acronym.

[5]https://github.com/aistairc/RDF-star2Vec

TABLE 1. Comparison table describing embedding methods for RDF and hyper-relational KGs.

| | RDF-star2Vec (Ours) | RDF2Vec [8] | RDF2Vec$_{oa}$ [23] | m-TransH [26] | NaLP [12] | StarE [11] |
|---|---|---|---|---|---|---|
| **Target tasks** | Data mining | | | Link prediction | | |
| **Types of KGE** | Graph walk | | | Translation | FCN | GNN |
| **Handling RDF data directly** | ✓ | ✓ | ✓ | - | - | - |
| **Handling RDF-star data directly** | ✓ | - | - | - | - | - |
| **Hyper-relational KGE** | ✓ | - | - | ✓ | ✓ | ✓ |
| **Recursive hyper-relational KGE** | ✓ | - | - | - | - | - |
| **QT embedding** | ✓ | - | - | - | - | - |
| **QT–QT relation embedding** | ✓ | - | - | - | - | - |

Steenwinckel et al. [25] showed that the WL graph kernel offers little improvements in the context of a single KG with respect to walk embeddings, and proposed five alternative walking strategies.

In this way, numerous KGE methods for RDF have been developed, and most of them were influenced by RDF2Vec. Our approach is positioned as an extension of RDF2Vec. To the best of our knowledge, our approach is the first method capable of directly representing recursive hyper-relational KGs described in RDF-star in low-dimensional vector space without any information loss.

### B. HYPER-RELATIONAL KNOWLEDGE GRAPH EMBEDDINGS

Wen et al. [26] proposed m-TransH, a generalized model of TransH [27] for link prediction of Hyper-relational KGs. However, m-TransH does not consider the relatedness of the components in the same hyper-relation. Guan et al. [12] proposed NaLP, a method to explicitly model the relatedness of the role-value pairs involved in the same hyper-relation. While m-TransH is grounded in a translation-based link prediction model, NaLP employs a fully connected neural network (FCN). Galkin et al. [11] proposed StarE, a graph neural network (GNN)-based link prediction model for hyper-relational KG. StarE extended CompGCN [28] to handle *qualifiers* of Wikidata. However, these approaches focus on link prediction in simple hyper-relational KGs without recursive structures. Therefore, these approaches are different from the purpose of this study. In addition, they are unable to load standard RDF or RDF-star data directly. Note that these studies have generated their benchmark datasets. A comparison table describing the embedding methods for RDF and hyper-relational KGs is presented in Table 1.

Kwan et al. [10] proposed ExtRet, an algorithm designed to convert RDF-star graphs into regular RDF graphs. ExtRet extended the representation of standard RDF Reification to minimize structural information loss to perform link predictions on binary relations. In contrast, our method can generate embeddings directly without converting the structure of the original RDF-star graphs. Therefore, no additional properties or intermediate nodes are generated due to the conversion, and the embeddings can be produced for the original RDF-star without any information loss.

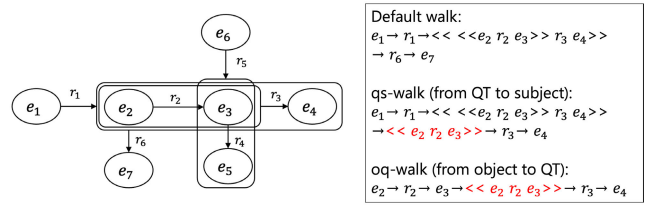

FIGURE 2. Graph walk methods for RDF-star.

### C. BENCHMARKING DATASETS FOR HYPER-RELATIONAL KNOWLEDGE GRAPH EMBEDDING

A well-known dataset in the realm of hyper-relational KG is Wikidata [29], which contains *qualifiers* to specify relationships and represents supplementary information in a key-value format. WikiPeople [12], a benchmark dataset consisting of information about individuals, is also extracted from Wikidata and serves the purpose of evaluating link prediction models for hyper-relational KGs. JF17K [26], another benchmark, is extracted from Freebase [30]. However, it was pointed out that WikiPeople contains numerous literal values that are conventionally ignored in KGE approaches, and JF17K has been criticized for significant test leakage. As a response, Galkin et al. [11] proposed WD50K as an alternative benchmark dataset designed for the link prediction tasks.

However, existing datasets lack complex structures such as multi-leveled nested QTs and QT–QT relations. Thus, we have developed a complex RDF-star dataset based on KGRC-RDF [15], [16], [31] and integrated its gold standard dataset into GEval [17], thereby enabling evaluations in classification, clustering, entity relatedness, and QT similarity tasks.

## III. RDF-STAR2VEC
### A. GRAPH WALKS
The proposed RDF-star2Vec is a graph walk-based embedding model based on RDF2Vec. Let $e \in \mathcal{E}$ denote an entity with Uniform Resource Identifier (URI) (i.e., the resource in RDF) and $r \in \mathcal{R}$ denotes a relation (i.e., the property in RDF). A QT $q \in \mathcal{Q}$ is represented by a triple of $q := (n, r, n)$, where node $n \in \mathcal{E} \cup \mathcal{Q}$. Therefore, RDF-star graph is $\mathcal{G} \subseteq (\mathcal{E} \cup \mathcal{Q}) \times \mathcal{R} \times (\mathcal{E} \cup \mathcal{Q})$.

**Algorithm 1** Random Walk of RDF-Star2Vec
**Require:** Root node $e$, # of walks $n$, Depth $d$, Map of QTs $qts$
**Ensure:** Walk list $wl$
1: $wl \leftarrow$ an empty list
2: **for** $currentDepth < d$ **do**
3:   **if** $currentDepth$ is 0 **then**
4:     $triples_e^{subj} \leftarrow$ getTriplesInvolvingSubj($e$)
5:     /* $triples_e^{subj}$ means a list of triples which involving $e$ as subject */
6:     $qt_e^{obj} \leftarrow$ getRandQtInvolvingObj($e$, $triples_e^{obj}$, $qts$)
7:     /* $qt_e^{obj}$ means a QT which involving $e$ as object */
8:     $qt_e^{subj} \leftarrow$ getRandQtInvolvingSubj($e$, $triples_e^{subj}$, $qts$)
9:     $walk \leftarrow$ an empty list
10:     /* Continue to Algorithm 2 */
11:     $wl \leftarrow$ QTWalk($wl$, $walk$, $triples_e^{subj}$, $qt_e^{obj}$, $qt_e^{subj}$)
12:   **else**
13:     **for** $walk \in wl$ **do**
14:       $lastObj \leftarrow walk$.get($walk$.size - 1)
15:       **if** $walk$ is not QT **then**
16:         $lastPred \leftarrow walk$.get($walk$.size - 2)
17:         $lastSubj \leftarrow walk$.get($walk$.size - 3)
18:       **end if**
19:       $triples_{lastObj}^{subj} \leftarrow$ getTriplesInvolvingSubj($lastObj$)
20:       $qt_{lastObj}^{obj} \leftarrow$ getQT($lastSubj$, $lastPred$, $lastObj$)
21:       $qt_{lastObj}^{subj} \leftarrow$ getRandQtInvolvingSubj($lastObj$, $triples_{lastObj}^{subj}$, $qts$)
22:       $triples_e^{subj} \leftarrow triples_{lastObj}^{subj}$
23:       $qt_e^{obj} \leftarrow qt_{lastObj}^{obj}$
24:       $qt_e^{subj} \leftarrow qt_{lastObj}^{subj}$
25:       /* Continue to Algorithm 2 */
26:       $wl \leftarrow$ QTWalk($wl$, $walk$, $triples_e^{subj}$, $qt_e^{obj}$, $qt_e^{subj}$)
27:     **end for**
28:   **end if**
29:   /* trim the list */
30:   **while** $wl$.size > $n$ **do**
31:     remove a random item from $wl$
32:   **end while**
33: **end for**

**Algorithm 2** QT-Walk Generation
**Require:** $wl$, $walk$, $triples_e^{subj}$, $qt_e^{obj}$, $qt_e^{subj}$, $\alpha$, $\beta$
**Ensure:** $wl$
1: $rand_{oq} \leftarrow$ a random number
2: $rand_{qs} \leftarrow$ a random number
3: $newWalk \leftarrow$ a copy of $walk$
4: **if** $qt_e^{obj}$ is not null AND $rand_{oq} < \beta$ **then**
5:   /* oq-walk: from object to QT */
6:   $qt \leftarrow qt_e^{obj}$.qt
7:   /* $qt_e^{obj}$.qt means $qt_e^{obj}$ with quotes, i.e., <<subj pred obj>> */
8:   **if** $walk$ is empty **then**
9:     append $qt_e^{obj}$.obj to $newWalk$
10:     /* $qt_e^{obj}$.obj means the object of the $qt_e^{obj}$ */
11:   **end if**
12:   append $qt$ to $newWalk$
13:   remove $walk$ from $wl$
14:   append $newWalk$ to $wl$
15: **else if** $qt_e^{subj}$ is not null AND $rand_{qs} < \alpha$ **then**
16:   /* qs-walk: from QT to subject */
17:   **if** $walk$ is empty **then**
18:     append $qt_e^{subj}$.qt to $newWalk$
19:   **end if**
20:   append $qt_e^{subj}$.subj to $newWalk$
21:   append $qt_e^{subj}$.pred to $newWalk$
22:   append $qt_e^{subj}$.obj to $newWalk$
23:   append $newWalk$ to $wl$
24: **else**
25:   /* default walk */
26:   remove $walk$ from $wl$
27:   **for** $triple \in triples_e^{subj}$ **do**
28:     append predicate of $triple$ to $newWalk$
29:     append object of $triple$ to $newWalk$
30:     append $newWalk$ to $wl$
31:   **end for**
32: **end if**

An example of an RDF-star graph containing the complex structure used in the following description is displayed in Figure 2. Similar to the default walks in RDF2Vec, the walk path is as $e_1 \rightarrow r_1 \rightarrow \ll\ll e_2\ r_2\ e_3 \gg r_3\ e_4 \gg \rightarrow r_6 \rightarrow e_7$, when starting from $e_1$, where $\ll\ll e_2\ r_2\ e_3 \gg r_3\ e_4 \gg$ denotes a multi-leveled nested QT. Similar to other entities, this multi-leveled nested QT is treated as a node. Thus, in the default walk, for each node $n \in \mathcal{E} \cup \mathcal{Q}$ in a given $\mathcal{G}$, we generate all sequences $S_n^d$ of depth $d$ rooted in the node $n$. The $S_n^d$ represents a set of sequences $s_n'^d$ expressed in Equation 1.

$$s_n'^d = n, r_{1,j}, n_{1,j}, r_{2,j}, n_{2,j}, \ldots, r_{d,j}, n_{d,j} \quad (1)$$

where $j \in \mathcal{R}(n)$ denotes an element of the preceding node's relationships. Thus, the final sequence set generated by the default walks at depth $d$ on the single RDF-star graph is $\bigcup_{n \in \mathcal{E} \cup \mathcal{Q}} S_n^d$.

As a limitation, the default walks cannot accurately extract the semantics originally expressed by QTs because it cannot walk the compositional entities of QTs. Thus, we propose a new walk method "QT-walk" between the QTs and their compositional entities. Specifically, we propose the following two types of QT-walk.

(1) **qs-walk**: walk from a QT to its compositional entity in the subject role
(2) **oq-walk**: walk from a compositional entity in the object role to the QT

For example, in Figure 2, the qs-walk generates a sequence as $e_1 \rightarrow r_1 \rightarrow \ll\ll e_2\ r_2\ e_3 \gg r_3\ e_4 \gg \rightarrow \ll e_2\ r_2\ e_3 \gg \rightarrow r_3 \rightarrow e_4$ when the starting point is $e_1$. The oq-walk generates a sequence as $e_2 \rightarrow r_2 \rightarrow e_3 \rightarrow \ll e_2\ r_2\ e_3 \gg \rightarrow r_3 \rightarrow e_4$ when the starting point is $e_2$. Thus, in the QT-walk, for a given $\mathcal{G}$, for each node $n \in \mathcal{E} \cup \mathcal{Q}$, we generate all sequence $P_n^d$ of depth $d$ rooted in the node $n$. The $P_n^d$ is a set of sequences $p_n'^d$ shown in Equation 2.

$$p'^d_n = n, w(n_{1,j}), w(n_{2,j}), \ldots, w(n_{d,j}) \quad (2)$$

where $w(n_{i,j})$ is a QT $q_{i,j}$, a triple $(n_{i,j}^{subj}, r_{i,j}, n_{i,j}^{obj})$, or a pair $(r_{i+1,j}, n_{i+1,j})$. The $q_{i,j}$ is an expression of a QT with quotes ("≪" and "≫"). The $(n_{i,j}^{subj}, r_{i,j}, n_{i,j}^{obj})$ represents a compositional triple of a QT. The $(r_{i+1,j}, n_{i+1,j})$ denotes the pair of predicate and object related to $n_{i,j}$. Therefore, the final set of the sequences generated by the QT-walk of depth $d$ on the single RDF-star graph is $\bigcup_{n \in \mathcal{E} \cup \mathcal{Q}} P^d_n$.

Algorithms 1 and 2 describe the graph walk of our method. Specifically, Algorithm 1 represents an algorithm that retrieves candidate QTs and triples for the graph walk. Algorithm 2 is the QT-walk that cases $w(n_{i,j})$ in Equation 2. In addition, we introduce two parameters $\alpha$ and $\beta$ to set the transition probability. The $\alpha$ is the transition probability from a QT to its compositional entity in the subject role. The $\beta$ is the transition probability from a compositional entity in the object role to the QT. The former walks deeply into the nested structure, while the latter walks out of the nested structure. The parameters $\alpha$ and $\beta$ are set manually in the range $\alpha \in (0, 1]$ and $\beta \in (0, 1]$, respectively. If both the qs-walk and oq-walk transitions are possible, priority is given to the oq-walk.

We also introduce mid walk [22] to RDF-star2Vec. Instead of starting random walks at all entities of interest, it is randomly decided for each depth iteration whether to go backward, i.e., to one of the node's predecessors, or forwards, i.e., to the node's successors. The mid walk of the proposed method is explained in Algorithm 3.

### B. REPRESENTATION LEARNING

DeepWalk [32], node2vec [33], RDF2Vec [8] and other representative graph walk-based KGE methods use word2vec [20], a neural network-based representation learning method of words, to encode entities and relations into distributed representations from the generated sequence set. Word2vec offers two principal learning algorithms: the continuous bag-of-words (CBOW) and skip-gram models. Our proposed method utilizes the skip-gram model for learning distributed representations from sequence sets, as empirical evidence suggests that the skip-gram model outperforms CBOW model in the context of RDF2Vec. The skip-gram model aims to maximize the average log probability denoted by Equation 3, given a sequence of training words $w_1, w_2, \ldots, w_t$.

$$\frac{1}{T} \sum_{t=1}^{T} \sum_{j} \log(p(w_{t+j}|w_t)) \quad (3)$$

where $T$ indicates the number of words and $w_{t+j}$ ($-c \leq j \leq c$) represents the words appearing in context window $c$. The probability $p(w_o|w_i)$ of an input word $w_i$ and an output word $w_o$ is calculated using the Softmax function in Equation 4.

$$p(w_o|w_i) = \frac{exp(v'^T_{w_o} \cdot v_{w_i})}{\sum_{w \in W} exp(v'^T_w \cdot v_{w_i})} \quad (4)$$

**Algorithm 3** Mid Walk of RDF-Star2Vec
**Require:** Node $e$, # of walks $n$, Depth $d$, Map of QTs $qts$
**Ensure:** Walks list $wl$
1: $wl \leftarrow$ an empty list
2: $np \leftarrow e$ /* next predecessor */
3: $ns \leftarrow e$ /* next successor */
4: **while** $wl$.size $< n$ **do**
5:    **while** $currentDepth < d$ **do**
6:      $rand_{oq} \leftarrow$ a random number
7:      $rand_{qs} \leftarrow$ a random number
8:      $zo \leftarrow$ randomPick({0, 1})
9:      **if** $zo$ is 0 **then**
10:        /* next predecessor */
11:        $tpls^{obj}_{np} \leftarrow$ getTriplesInvolvingObj($np$)
12:        $qt^{obj}_{np} \leftarrow$ getRandQtInvolvingObj($np$, $tpls^{obj}_{np}$, $qts$)
13:        **if** $qt^{obj}_{np}$ is not null AND $rand_{oq} < \beta$ **then**
14:          /* oq-walk */
15:          append $qt^{obj}_{np}$.obj to the beginning of $walk$
16:          append $qt^{obj}_{np}$.pred to the beginning of $walk$
17:          append $qt^{obj}_{np}$.subj to the beginning of $walk$
18:          currentDepth++
19:          $np \leftarrow qt^{obj}_{np}$.subj
20:        **else**
21:          $triple \leftarrow$ getRandomTriple($tpls^{obj}_{np}$)
22:          append $triple$ to the beginning of $walk$
23:          $np \leftarrow$ the subject of $triple$
24:        **end if**
25:      **else**
26:        /* next successor */
27:        **if** $walk$.size >= 3 **then**
28:          $lastPred \leftarrow walk$.get($walk$.size - 2)
29:          $lastSubj \leftarrow walk$.get($walk$.size - 3)
30:        **end if**
31:        $tpls^{subj}_{ns} \leftarrow$ gettplsInvolvingSubj($ns$)
32:        $qt^{subj}_{ns} \leftarrow$ getRandQtInvolvingSubj($ns$, $tpls^{subj}_{ns}$, $qts$)
33:        **if** $qt^{subj}_{ns}$ is not null AND $rand_{qs} < \alpha$ **then**
34:          /* qs-walk */
35:          append $qt^{subj}_{ns}$.subj to the end of $walk$
36:          append $qt^{subj}_{ns}$.pred to the end of $walk$
37:          append $qt^{subj}_{ns}$.obj to the end $walk$
38:          $ns \leftarrow qt^{subj}_{ns}$.obj
39:        **else**
40:          $triple \leftarrow$ getRandomTriple($tpls^{subj}_{ns}$)
41:          append $triple$ to the beginning of $walk$
42:          $ns \leftarrow$ the subject of $triple$
43:        **end if**
44:      **end if**
45:    **end while**
46:    append $walk$ to $wl$
47: **end while**

where $v_w$ and $v'_w$ represent the input and output vectors of the word $w$, and $W$ is the word set.

Portisch et al. demonstrated a partial enhancement in RDF2Vec's performance by incorporating structured word2vec [14], which takes word order into account [23]. Therefore, in the proposed method, we employ structured word2vec, specifically the structured skip-gram model, expecting to improve the embedding performance of the context of the QT-walk. Figure 3(b) illustrates the architecture of the structured skip-gram. The classic skip-gram uses a single output matrix $O \in \Re^{|W| \times d}$ ($d$=dimension) to predict every contextual word $w_{i-c}, \ldots, w_{i-1}, w_{i+1}, \ldots, w_{i+c}$, given the embeddings of the center word $w_i$; in contrast, the structured skip-gram uses different output matrix $O_{o-i}$ for a specific relative position to the center word.

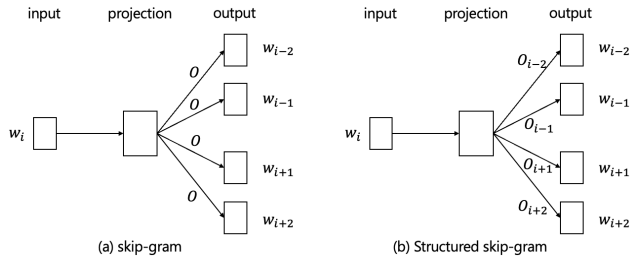

FIGURE 3. Architecture of the skip-gram [20] and structured skip-gram model [14].

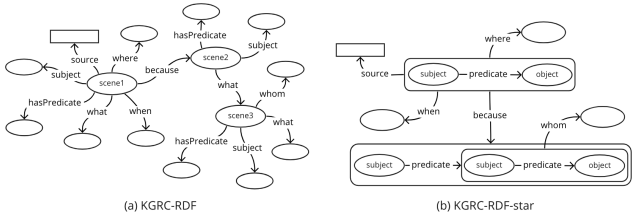

FIGURE 4. Data structures of KGRC-RDF and KGRC-RDF-star.

## IV. DATASET

Given that existing hyper-relational KG datasets such as WikiPeople, JF17K, and WD50K lack complex structures, they are unsuitable for evaluating the proposed method. Thus, we provide a dataset KGRC-RDF-star that is available for benchmarking the embeddings of complex RDF-star graphs by converting the KGRC-RDF[6] [15], [31] to RDF-star data, which is provided for International Knowledge Graph Reasoning Challenge (IKGRC).[7]

The original KGRC-RDF is a set of eight KGs built based on Sherlock Holmes mystery stories and has been published as an RDF dataset for benchmarking XAI systems that can provide reasons for its decisions. Figure 4(a) shows the data structure of KGRC-RDF. The KGRC-RDF contains detailed descriptions of "who, what, when, where, why, and how (5W1H)" information along with the storyline and has a rich variation of entities as values of 5W1H. Several approaches [31], [34], [35] were proposed to estimate the criminal from the KGRC-RDF and present a convincing explanation by KGE and logical reasoning technologies.

We converted the KGRC-RDF graphs to the RDF-star graphs, as shown in Figure 4(b), and provided it as a dataset for evaluating the RDF-star embeddings. For example, the scene "Julia met a lieutenant commander two years ago at Harrow." was described in the KGRC-RDF as follows:

[6]https://github.com/KnowledgeGraphJapan/KGRC-RDF/tree/ikgrc2023
[7]https://ikgrc.org/

TABLE 2. Statistics of the KGRC-RDF-star.

| | Class | 17 |
|---|---|---|
| | Instance | 9,524 |
| | Property | 640 |
| Triple | Standard triple | 14,180 |
| | Single-nested QT | 9,765 |
| | Double-nested QT | 6,409 |
| | Triple-nested QT | 695 |
| | Quadruple-nested QT | 43 |
| | Total | 31,092 |

```
1  @prefix kgc: <http://kgc.knowledge-
       graph.jp/ontology/kgc.owl#> .
2  @prefix kdsb: <http://kgc.knowledge-
       graph.jp/data/SpeckledBand/> .
3  kdsb:36 a kgc:Situation ;
4  kgc:hasPredicate kdp:meet ;
5  kgc:subject kdsb:Julia ;
6  kgc:then kdsb:37 ;
7  kgc:when kdsb:2_years_ago ;
```

The above description was converted as follows.

```
1  << kdsb:Julia kdp:meet kdsb:
       lieutenant_commander >>
2    a   kgc:Situation ;
3  kgc:then   << kdsb:Julia kdp:engage
         kdsb:lieutenant_commander >> ;
4  kgc:when   kdsb:2_years_ago ;
5  kgc:where  kdsb:Harrow .
```

It is necessary to specify an object of a triple explicitly when converting from the KGRC-RDF to the KGRC-RDF-star. We set the priority for the object selection as follows: *what > whom > where > on > to > from*. If either subject or object did not exist, owl:Nothing was substituted.

QTs are unique for each combination of subject $s$, predicate $p$, and object $o$, and there is no URI and ID for identifying a QT. However, the same $s$, $p$, and $o$ combinations might occur in different scenes when the KGRC-RDF is converted to the KGRC-RDF-star. It is necessary to distinguish these QTs and assign different metadata to them. Therefore, we solved this issue by assigning a unique ID to each QT and nested these triples as a QT as follows: $<< <<s\ p\ o >>\ id\ val >>\ p'\ o'$.

Table 2 provides a statistical overview of the KGRC-RDF-star. Notably, these statistics exclude the nesting introduced to address the issue of QT uniqueness. The finalized dataset has been made publicly accessible via GitHub.[8]

## V. EVALUATION
### A. EVALUATION TASKS

For the evaluation of RDF-star embeddings generated through our proposed method, we focus on classification,

[8]https://github.com/aistairc/KGRC-RDF-star

**TABLE 3.** Details of the classification task.

| | Dataset | Semantics of classes | Classes | Size |
|---|---|---|---|---|
| **INPUT** | PersonObjectPlace | Type of entities | 3 | 540 |
| | QT900 | Type of quoted triples | 3 | 900 |

| | Model | Configuration |
|---|---|---|
| **Model** | Naive Bayes | - |
| | C4.5 decision tree | - |
| | k-NN | k=3 |
| | SVM | $C \in \{10^{-3}, 10^{-2}, 0.1, 1, 10, 10^2, 10^3\}$ |

| | Metric | Range | Optimum |
|---|---|---|---|
| **OUTPUT** | Accuracy | [0,1] | Highest |

clustering, entity relatedness, and QT similarity tasks. We selected RDF2Vec and RDF2vec$_{oa}$ as the baseline methods of KGE. Here, we extended jRDF2Vec[9] to enable to generate sequences of RDF-star data, since RDF2vec cannot load RDF-star data. We employ an evaluation framework for graph embedding, GEval [17], to evaluate the performance of each task. Although the GEval provides the entity list of DBpedia [36] as the default gold standard dataset, DBpedia is not RDF-star data and is not appropriate for the evaluation in this experiment. Thus, we constructed gold standard datasets based on KGRC-RDF-star to support benchmarking of the complex RDF-star embeddings and incorporated them into the GEval, and then published them on GitHub.[10] In the following sections, we describe the details of each task and how to create gold standard datasets.

#### 1) CLASSIFICATION
The classification task learns training data for entity-label pairs and estimates the labels given unknown entities. Table 5 shows the details of the classification task and the gold standard datasets. These gold standard datasets have been designed to quantitatively evaluate whether each entity and QT has represented suitable feature vectors that contribute to the classification task. The dataset PersonObjectPlace comprises pairs of entities and labels, with labels classified into *Person*, *Object*, and *Place*. It is an unbiased dataset randomly extracted from the KGRC-RDF-star using SPARQL queries and designed to have an equal number of each class. Similarly, QT900 contains pairs of QTs and labels, which are classified into *Situation*, *Statement*, and *Thought*, and it is also an unbiased dataset designed to have an equal number of each class. In the classification task, the above values of the types are removed from the source RDF-star dataset when generating embeddings to exclude the potential ground truth information from the feature vectors. Finally, the results are calculated using 10-fold cross-validation.

**TABLE 4.** Details of the clustering task.

| | Dataset |
|---|---|
| **INPUT** | same as the Classification task |

| | Model | Configuration |
|---|---|---|
| **Model** | Agglomerative Clustering | similarity_metric |
| | Ward Hierarchical Clustering | similarity_metric |
| | DBscan | similarity_metric |
| | k-Means | - |

| | Metric | Range | Optimum |
|---|---|---|---|
| **OUTPUT** | adjusted rand score | [−1,1] | Highest |
| | adjusted mutual info score | [0,1] | Highest |
| | Fowlkes Mallow index | [0,1] | Highest |
| | v_measure score | [0,1] | Highest |
| | homogeneity score | [0,1] | Highest |
| | completeness score | [0,1] | Highest |
| | accuracy | [0,1] | Highest |

#### 2) CLUSTERING
In the clustering task, clusters are generated from the embeddings using unsupervised methods, and performance is evaluated by comparing the clusters to the gold standard dataset. Table 4 shows the details of the clustering task. The gold standard datasets are the same ones used in the classification task.

#### 3) ENTITY RELATEDNESS
In the entity relatedness task, we assume that two entities are related if they often appear in the same context, as in prior studies [9], [17]. Table 5 shows the details of the entity relatedness task. The kgrc_entity_relatedness is a set of 21 person entities extracted from the KGRC-RDF-star and 10 entities related to each person, and the related entities are sorted by their relatedness. This gold standard dataset was created as follows, referring to the methodology of KORE [37].

---
[9]https://github.com/dwslab/jRDF2Vec
[10]https://github.com/aistairc/GEval-forKGRC-RDF-star/

**TABLE 5.** Details of the entity relatedness task.

| | Dataset | Structure | Size |
|---|---|---|---|
| INPUT | kgrc_entity_relatedness | Person entity with a sorted list of 10 related entities | 210 |
| | **Model** | | **Config** |
| Model | Create sorted lists of similarity scores between each person entity and related entities | | similarity_metric |
| | **Metric** | **Range** | **Interpretation** |
| OUTPUT | Kendall's Tau (Kendall Rank Correlation Coefficient) | $[-1,1]$ | Values close to 1: correlation Values close to 0: no correlation |

(1) Twenty-one person entities are extracted as seed entities from the KGRC-RDF-star.
(2) Ten candidates are extracted from sorted lists of entities that co-occur with the seed entity in the same scenes (210 in total).
(3) All possible comparisons of the 10 candidates with respect to their seed are created (945 in total).
(4) Amazon Mechanical Turk (MTurk)[11] workers are asked which of the given two entities is more related to the seed entity. Ten workers answer each question.
(5) Ten comparison pairs are created as the gold standard by authors to improve the quality of the crowdsourcing task. Spam workers are removed based on the answers for these pairs.
(6) All the comparisons are aggregated into a single confidence that one entity is more (or equally) related to the seed.
(7) The ten candidate entities are subsequently ranked according to these confidence values.

The evaluation model is a simple algorithm that sorts the list of candidate entities based on the similarity scores between the vectors of the seed entity and the candidate entities. Finally, the output list is compared to the gold standard dataset using Kendall's rank correlation coefficient.

### 4) QT SIMILARITY

A QT in the KGRC-RDF-star corresponds to a scene in KGRC-RDF. We assume that the vectors of similar QTs are located nearby in the embedding space by adequately learning the semantic relations between QTs' neighbors and between the compositional entities of the QTs. Thus, this task aims to verify whether the embedding vectors of the QTs contain semantic and contextual similarity equivalent to human judgment. Table 6 shows the details of the QT similarity task. QT50 was created through the following procedures, using the methodology of LP50 [38].

(1) Fifty QTs and the scene descriptions corresponding to the QTs are extracted from KGRC-RDF-star and KGRC.

[11]https://www.mturk.com/

(2) All possible pairs of the 50 QTs are created (1,225 in total).
(3) MTurk workers judge the similarity on a five-point scale (with one indicating "not similar at all" and five indicating "quite similar"). 10 workers answer each question.
(4) Five pairs are created as the gold standard by authors to improve the quality of the crowdsourcing task. Spam workers are removed based on the answers for these pairs.
(5) The average similarity scores of each pair are calculated.

### B. EVALUATION RESULTS

Table 7 shows the evaluation results. We used cosine similarity as a similarity metric. We preliminarily compared random walk and mid walk in RDF-star2Vec and found that the mid walk performed better than the random walk. Therefore, we adopted the mid walk in all of our evaluation experiments. We set the following hyperparameters: depth is 8, number of walks per entity is 100, window size is 5, dimension is 100, $\alpha$ is 0.5, and $\beta$ is 0.5. The proposed method outperformed RDF2Vec and RDF2Vec$_{oa}$ on more than half of the tasks. Specifically, RDF-star2Vec with the structured skip-gram $(oa)$ achieved the highest accuracy in the classification task of the PersonObjectPlace dataset. Conversely, all methods achieved good performance in the classification task of the QT900 dataset, and RDF2Vec$_{oa}$ slightly outperformed RDF-star2Vec. This difference is slight enough to require a statistical test. To evaluate the significance between the baseline and RDF-star2Vec$_{oa}$, we performed a Wilcoxon signed rank test based on the results of 10-fold cross-validation using the classification model with the best accuracy for each method (i.e., Support Vector Machine (SVM)). We used a standard significance level of $p < 0.05$. The result showed that the RDF-star2Vec$_{oa}$ is significantly more accurate (p-value = 0.00758) than the baseline in the classification task of the PersonObjectPlace dataset. Conversely, the baseline outperformed the RDF-star2Vec$_{oa}$ by 0.002 for the classification of the QT900 dataset. However, a Wilcoxon signed rank test at a significance level of $p < 0.05$ revealed no significant difference in accuracy in regarding the classification task of the QT900 between the baseline and the RDF-star2Vec$_{oa}$ (p-value = 0.131).

**TABLE 6.** Details of the QT similarity task.

| | | | |
|---|---|---|---|
| **INPUT** | Dataset | Structure | Size |
| | QT50 | QT1 QT2 avg | 50 QTs |
| **Model** | Model | Configuration | |
| | similarity(QT1, QT2) | similarity_metric | |
| **OUTPUT** | Metric | Range | Interpretation |
| | Pearson correlation (P_cor) | [-1,1] | Close to 1: correlation / Close to 0: no correlation |
| | Spearman correlation (S_cor) | [-1,1] | Close to 1: correlation / Close to 0: no correlation |
| | Harmonic mean of P_cor and S_cor | [-1,1] | Close to 1: correlation / Close to 0: no correlation |

**TABLE 7.** Evaluation results (parameter settings: depth=8, walks_per_entity=100, window=5, dimension=100, $\alpha$=0.5, $\beta$=0.5).

| Task | Metric | Dataset | RDF2vec [8] classic | RDF2vec [8] oa [23] | RDF-star2Vec classic | RDF-star2Vec oa |
|---|---|---|---|---|---|---|
| Classification | Accuracy | PersonObjectPlace | 0.744 | 0.800 | 0.759 | **0.822** |
| | Accuracy | QT900 | 0.968 | **0.976** | 0.971 | 0.974 |
| Clustering | Accuracy | PersonObjectPlace | 0.557 | 0.751 | 0.477 | **0.870** |
| | Accuracy | QT900 | **0.708** | 0.690 | 0.435 | 0.492 |
| Entity Relatedness | Kendall's Tau | kgrc_entity_relatedness | 0.422 | 0.466 | **0.600** | 0.555 |
| QT Similarity | P_cor | QT50 | 0.0325 | 0.0299 | **0.116** | 0.100 |

RDF-star2Vec$_{oa}$ achieved the highest accuracy in the PersonObjectPlace dataset in the clustering task. Conversely, the classic RDF2Vec achieved the highest accuracy in the QT900 dataset. Our proposed method is inclined to walk from a QT to its compositional entities rather than to the rdf:type of the QT. In contrast, RDF2Vec frequently walks to the rdf:type, which corresponds to the cluster information in the gold standard dataset. As a result, the clusters generated by the proposed method deviated from those of the gold standard dataset.

In the entity relatedness task, we observed the highest correlation when using RDF-star2Vec with the classic skip-gram ($p = 0.0166$; $p < 0.05$ was taken to indicate statistical significance). The embedding of the proposed method reflects entity relatedness more than the existing methods because the proposed method can place a QT and its compositional entities close to each other in the generated sequences. For instance, when two individuals appear in the same context, they serve as the QT's compositional subject or object entities and share metadata such as time, place, reason, and related scenes. Therefore, the proposed method was able to place these entities (i.e., entities of N-ary relations represented by RDF-star) close to each other in the generated sequence. Consequently, the skip-gram was able to embed entities considering the entity relatedness.

In the QT similarity task, none of the methods performed well. However, the results of the proposed method very slightly correlated with the gold standard dataset.

These results indicate that our proposed method is especially suitable for the classification and clustering tasks of normal entities in RDF-star datasets.

To summarize, RDF-star2Vec$_{oa}$ significantly outperformed the baseline in four of the six tasks, was inferior in one task, and equaled the baseline in another. Therefore, when considering all tasks, RDF-star2Vec$_{oa}$ overall outperforms the baseline.

### C. VISUALIZATION

We applied the t-SNE [39] to the 100-dimensional embedding vectors to visualize the results as a two-dimensional plot. Figure 5 shows a comparison between the embedding vectors of baseline (RDF2Vec$_{oa}$) and RDF-star2Vec$_{oa}$. We used the same hyperparameters as in the evaluation experiments in Section V-B.

Figure 5 (a) shows the clustering results using the DBSCAN [40] after reducing the 100-dimensional embedding vectors to two dimensions using t-SNE [39]. DBSCAN, a density-based clustering algorithm, automatically determines the number of clusters. In this visualization, we configured the parameters as follows: the neighborhood distance is 60, and the minimum number of samples in a neighborhood for a core point is 6. In the RDF2Vec$_{oa}$, 20 clusters were formed with finely distributed nodes. In addition, some clusters of the same color were scattered in different locations (e.g., light blue). Conversely, in the RDF-star2Vec$_{oa}$, 17 clusters were formed with fewer instances of noise. Therefore, the RDF-star2Vec's embeddings are better for visualizing cluster characteristics than the RDF2Vec.

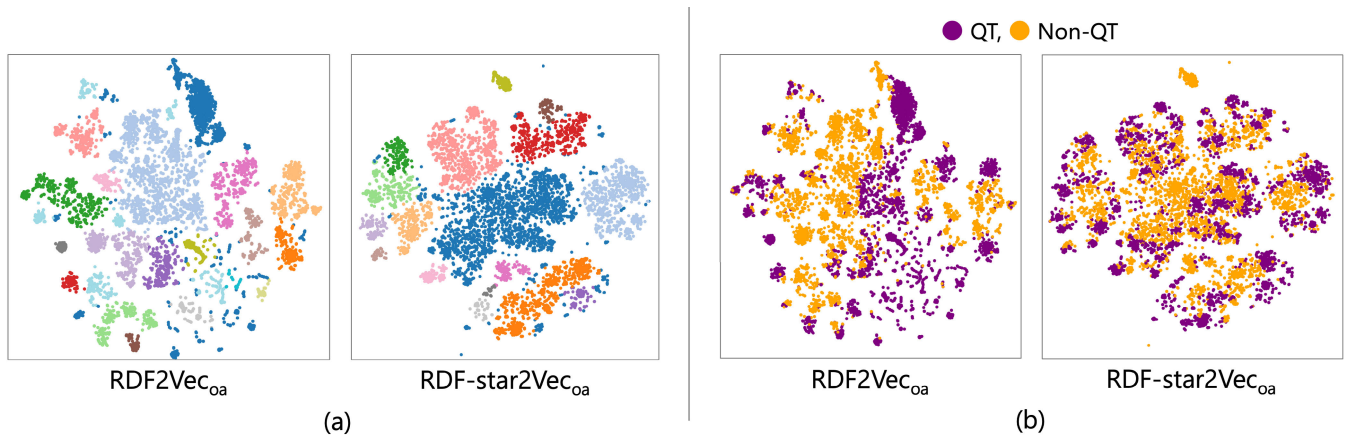

**FIGURE 5.** Visualization results of embeddings of baseline (RDF2Vec$_{oa}$) and RDF-star2Vec: (a) is the result of clustering using DBSCAN for 100-dimensional embeddings of all nodes, compressed to two dimensions using t-SNE, and (b) is color-coded to distinguish QTs from others.

Figure 5 (b) shows the embedding vectors, color-coded to distinguish QTs from other nodes. In the RDF2Vec$_{oa}$, there is a polarization between QT and others. Since the RDF2Vec cannot walk the entities constituting a QT, the internal semantics of the QT are not reflected in the embedding results. This polarization hinders any meaningful analysis of the relationships between QTs and normal entities. In contrast, RDF-star2Vec$_{oa}$ demonstrates no such polarization; it allows for the formation of mixed clusters containing both QTs and normal entities. The RDF-star2Vec can visualize QTs and normal entities closely since the internal semantics of the QTs are reflected in the embedding results. Therefore, from this visualization results, we can analyze the relationships between QTs and normal entities as well as the meaningful clusters they form.

### D. PARAMETER ANALYSIS AND DISCUSSION

We conducted more experiments by changing parameters to analyze and discuss the characteristics of RDF-star2Vec. Figure 6(a) shows the accuracy of the classification task for the PersonObjectPlace dataset for each combination of parameters $\alpha$ (i.e., probability of qs-walk) and $\beta$ (i.e., probability of oq-walk). Here, the structured skip-gram was used for representation learning. The results show that $\alpha = 0.2$ and $\beta = 0.2$ are the optimal combination, and $\alpha = 1.0$ and $\beta = 1.0$ are the worst combination. Since qs-walk and oq-walk are always performed when $\alpha = 1.0$ and $\beta = 1.0$, walks that loop on the same QT frequently occurs as follows: $q_1 \rightarrow e_1 \rightarrow r_1 \rightarrow e_2 \rightarrow q_1 \rightarrow e_1 \rightarrow r_1 \rightarrow \ldots \rightarrow q_1$. Therefore, it is difficult to adequately learn the representations of entities by using skip-gram because the variety of the generated sequences is reduced.

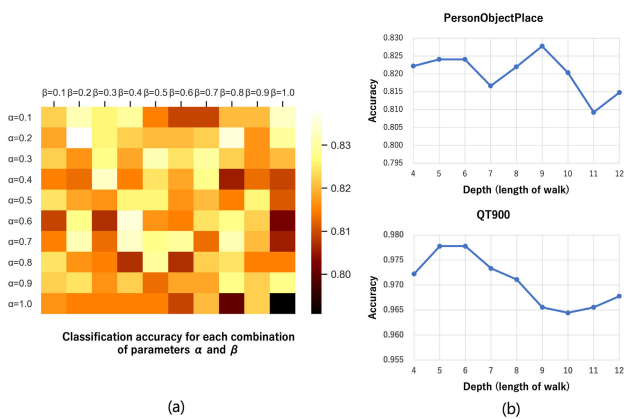

**FIGURE 6.** Parameter analysis for classification based on the RDF-star2Vec embeddings.

If $\alpha = 0$ and $\beta = 0$, it is equal to RDF2Vec,[12] which generates the sequence described in Equation 1. In KGRC-RDF-star, the variety of neighbors of normal entities is limited since the RDF2Vec's default walking strategy transition from QT to only metadata, such as "where," "when," "whom," "then," and "because." In contrast, our proposed method probabilistically transitions between QTs and their compositional entities. Hence, the proposed method can probabilistically place QTs before or after the normal entities in the generated sequences. Consequently, we observed that our proposed method improved the embedding performance as we assumed. We also found that values of $\alpha$ and $\beta$ less than or equal to 0.5 are suitable since excessive QT-walk leads to poor performance.

Figure 6(b) shows the accuracy of the classification tasks when changing the depth of the walks ($\alpha = 0.5$, $\beta = 0.5$). The figure shows that the large depth of the walks decreases accuracy. The deeper the walks in RDF-star2Vec, the greater the number of transitions between QTs and their compositional entities, and the loops will occur as described above. Therefore, such excessive QT-walk affected the accuracy.

---
[12]Note that this is an extended version of the original RDF2Vec to walk on RDF-star graphs but does not use QT-walk.

Several issues remain to be addressed to improve embedding performance. For example, qs-walk and oq-walk implemented in this study can be further classified in detail. Specifically, the qs-walk transitions from a QT to its compositional subject entity and then walk to the object in the same QT since we emphasize the context of the QT-walk. However, in KGRC-RDF-star, it is also possible to walk from this subject entity to other entities that are not components of the QT. We can also consider another version of oq-walk that ignores the context. Although we did not implement such walks that ignore the context of the QT-walk, it is possible to break away from the loop of the QT-walk using them. In addition, it is also possible to implement qo-walk, which transitions from a QT to its compositional object entity, and sq-walk, which transitions from a subject entity to a QT. In future work, we will implement new parameters to allow these walks. Furthermore, we will incorporate other walk flavors [24], [25].

### E. LIMITATIONS AND POTENTIAL BIASES OF DATASETS

We constructed KGRC-RDF-star based on KGRC-RDF. The KGRC-RDF consists of eight KGs constructed based on eight mystery stories. Although the mystery stories are closed worlds, they are datasets rich in diversity, with a wide variety of entities such as characters, animals, places, objects, events, and statements, and containing various relationships such as actions, types, person relationships, and causal relationships. Thus, we consider the KGRC-RDF and KGRC-RDF-star to be generalized datasets compared to domain-specific datasets such as biomedical KGs.

According to Kozaki et al. [16], there is a bias in the number of some properties among the eight KGs in KGRC-RDF. Since KGRC-RDF-star is based on these KGs, there is also a bias in the number of some properties among the novels. However, this bias does not directly affect the labels in our gold standard dataset.

For instance, our PersonObjectPlace dataset consists of 540 entities with cluster labels based on its rdf:type information (# of Person=180, # of Object=180, # of Place=180). Similarly, the QT900 dataset consists of 900 QTs with cluster labels based on its rdf:type information (# of Scene=300, # of Situation=300, # of Thought=300). Therefore, there is no bias in the number of correct labels. However, the variation in the novels from which the gold standard data was extracted is slightly biased. This is because there is a bias in the number of characters and places in different novels. In other words, each eight KG has its own characteristics, and we consider that evaluation experiments can be conducted according to the purpose by restricting the number of KGs used in an experiment.

The experimental results are somewhat generalized since this study used the eight KGs. We expect the future development of an even more robust benchmark dataset by including additional data.

## VI. CONCLUSION

In this paper, we proposed RDF-star2Vec, a novel graph walk-based KGE method, which directly represents RDF-star graphs with complex structures, such as multi-leveled nested QTs and QT–QT relations, in low-dimensional vector space without any information loss. The approach overcomes the challenge of embedding RDF-star's QTs in vector space by placing related entities closer together in generated sequences.

In addition, we constructed an RDF-star dataset containing complex structures based on KGRC-RDF, and incorporated four gold standard datasets based on it into GEval to enable benchmarking RDF-star embeddings. This framework facilitates evaluating the performance of future KGE methods for RDF-star.

Our proposed method demonstrated significantly better performance than existing methods in tasks such as classification, clustering, entity relatedness, and QT similarity tasks. Future work includes implementing other possible walks, optimizing hyperparameters, and constructing an RDF-star dataset to evaluate additional tasks such as regression, semantic analogy, and link prediction.

It is expected that many N-ary relation graphs (i.e., Hyper-relational KGs) will be published in RDF-star format in the future. We believe that the proposed method will be used as data mining tasks for RDF-star in various domains in the future.

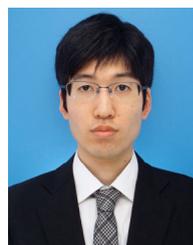

**SHUSAKU EGAMI** received the Ph.D. degree in engineering from The University of Electro-Communications, Tokyo, Japan, in 2019. He is currently a Senior Researcher with the National Institute of Advanced Industrial Science and Technology, Japan. He is also a part-time Lecturer with Hosei University, Tokyo, and a Collaborative Associate Professor with The University of Electro-Communications. His research interests include the semantic web, ontologies, and knowledge graphs.

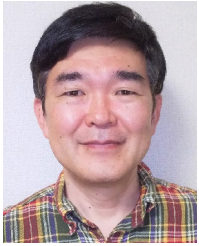

**TAKANORI UGAI** received the Ph.D. degree in engineering from the Tokyo Institute of Technology, Tokyo, Japan, in 2013. Since 1992, he has been a Researcher with Fujitsu Ltd., Japan. He is also a Researcher with the National Institute of Advanced Industrial Science and Technology, Japan, and a Lecturer with Tsukuba University, Ibaraki, Japan. His research interests include knowledge graphs, graph neural networks, and requirements engineering.

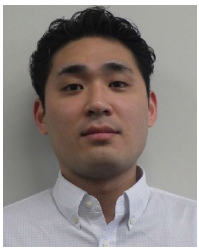

**MASATERU OOTA** received the master's degree from the Department of Mathematics, Hokkaido University, Japan, in 2015, and the Ph.D. degree in informatics with the Graduate School of Informatics and Engineering, The University of Electro-Communications, Tokyo, Japan, in 2023. He is a System Engineer at the National Institute of Advanced Industrial Science and Technology, Japan. His research interests include knowledge graphs and scene graph generation.

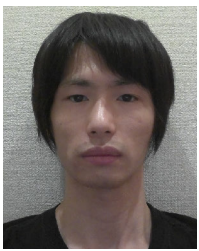

**KYOUMOTO MATSUSHITA** received the master's degree in computer science from the Kyushu Institute of Technology, Japan, in 2018. He is currently a Researcher with Fujitsu Ltd., Japan.

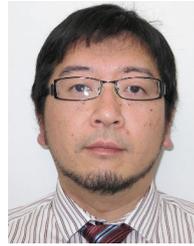

**TAKAHIRO KAWAMURA** received the M.S. and Ph.D. degrees in electrical engineering from Waseda University. From 2001 to 2002, he was a Visiting Scholar with The Robotics Institute, Carnegie Mellon University. He is currently the Deputy Director of the Research Center for Agricultural Information Technology, National Agriculture and Food Research Organization, Japan. His current research interest includes knowledge graph construction and its analysis combining semantic technology and machine learning. He was a Board Member of the Japanese Society for Artificial Intelligence (JSAI). He has served as the Chair for SIG on Semantic Web and Ontology in JSAI and a member of the Program/Organizing Committees of numerous conferences and workshops, including the International Semantic Web Conference.

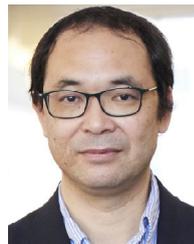

**KOUJI KOZAKI** received the Ph.D. degree in engineering from Osaka University, in 2002. He is currently a Professor of Osaka Electro-Communication University (OECU), Osaka, Japan. His field of expertise is computer science, in particular, Ontology Engineering and Semantic Technologies. A major theme of the research is the development of Hozo, which is an environment for building/using ontologies, and ontology building in several domains, such as clinical medicine, bioinformatics, and environmental engineering. His research interests include ontology building, fundamental theory of ontological engineering (especially for role and identity), ontology development systems, and ontology-based application. He was the Co-Program Committee Chair of the Joint International Semantic Technology Conference (JIST), in 2015, and the Co-General Chair of JIST2018.

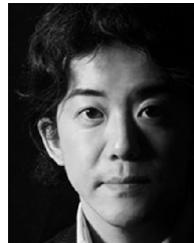

**KEN FUKUDA** received the Ph.D. degree in information science from The University of Tokyo, in 2001. In 2001, he joined the National Institute of Advanced Industrial Science and Technology (AIST) as a Research Scientist. He was a Visiting Lecturer with The University of Tokyo and a Visiting Associate Professor with Waseda University. He is currently leading the Data Knowledge Integration Team, Artificial Intelligence Research Center, AIST. Since then, he has led multiple national and international projects, covering a broad range of interdisciplinary research from life science to service science. His research interests include knowledge representation, knowledge graphs, data-knowledge integration, and human–robot interaction. He was awarded a Research Fellowship for Young Scientists as a Ph.D. Student.

○ ○ ○